# Intelligent Exploration of Solution Spaces Exemplified by Industrial Reconfiguration Management

Timo Müller [a,b]*, Benjamin Maschler [a,b], Daniel Dittler [a], Nasser Jazdi [a], Michael Weyrich [a]

[a] University of Stuttgart, Institute of Industrial Automation and Software Engineering, Pfaffenwaldring 47, 70569 Stuttgart, Germany
[b] These authors contributed equally to this publication.

* Corresponding author. Tel.: +49 711 685 67292; E-mail address: timo.mueller@ias.uni-stuttgart.de

**Abstract**

Many decision-making approaches rely on the exploration of solution spaces with regards to specified criteria. However, in complex environments, brute-force exploration strategies are usually not feasible. As an alternative, we propose the combination of an exploration task's vertical sub-division into layers representing different sequentially interdependent sub-problems of the paramount problem and a horizontal sub-division into self-sustained solution sub-spaces. In this paper, we present a universal methodology for the intelligent exploration of solution spaces and derive a use-case specific example from the field of reconfiguration management in industry 4.0.

## 1. Introduction

Concerning decision making for manifold problems, humans historically rely on already known solutions based on experience, often not following objective criteria. Nowadays, multiple (or: multi) criteria decision making (MCDM), i.e. the use of numeric techniques to consider quantified trade-offs between criteria of inherently different dimensions, is a topic of great and ever-growing importance. This is caused by the complex nature of many of today's problems, which cannot be solved using a single criterion, e.g. money, but need to satisfy requirements on a more complex plane, e.g. also involving non-monetary social or environmental criteria [1, 2].

This importance is reflected in the great number of scientific publications covering MCDM methods in various domains, from e.g. automotive supply chains [3] to sustainable energy generation [4, 5]. However, many studies focus only on the later stages of MCDM, taking the solution space as a given [6]. Yet, an intelligent exploration of solution spaces offers great potential to reduce complexity and computational burden of any MCDM process – and other complex optimization applications.

This becomes obvious on the example of reconfiguration management for discrete manufacturing, where frequent alterations to products as well as production processes and resources together with many parameters call for methods apart from brute forcing through all conceivable combinations.

*Objective*: In this article, a universal methodology for the intelligent exploration of solution spaces is presented and practical advice regarding its implementation given. It is then exemplarily evaluated on a reconfiguration management use case from the discrete manufacturing domain.

*Structure*: In chapter 2, a short overview of MCDM methods in different domains is presented. Then, the universal methodology for intelligent solution space exploration is introduced in chapter 3. Chapter 4 describes the methodology's evaluation on a discrete manufacturing use case. Finally, chapter 5 provides a conclusion and an outlook.

## 2. Related Work

MCDM procedures can be classified according to their solution space. Procedures operating within a discrete solution space are classified as Multiple Attribute Decision Making (MADM) and procedures operating within a continuous solution space as Multiple Objective Decision Making (MODM) [7]. In this context, classical MADM methods for finding the optimal solution from exponentiated solution variants are utility analysis, analytic hierarchy process or analytic network process [8]. MODM methods solve the (infinite) solution spaces with the help of mathematical algorithms [9]. In the following, some examples of the widespread use of MCDM methods in different domains is given.

In [4], an extensive systematic literature review on different MCDM approaches for renewable energy site selection is conducted under special consideration of the five site selection stages. It observes a strong increase in the use of MCDM methods for site selection. Covering a similar topic, [5] presents an overview of different MCDM approaches, specifically MADM methods, for sustainability assessment of grid-connected energy systems and shows that these methods use different criteria and objectives for decision support.

The authors in [3] review sustainability concepts in the automotive sector and show that most established concepts rarely consider trade-offs between environmental, social, and economic criteria. In their study, they go on to apply MCDM to consider management values as well as material performance properties as decision criteria.

In [1], the authors show that MCDM approaches have gained importance in recent years, especially in solving production, distribution as well as inventory decisions, and explore the use of different approaches for green supply chains. Sustainability dimensions also play an important role in [10], where the authors give an overview of MCDM approaches for sustainable production and show that so far, mainly fuzzy-based single model approaches are used here.

In [11], the authors describe a process planning approach for reconfigurable manufacturing systems. To solve their multi-objective evaluation, they utilize MODM methods. The authors derive that quality is an important factor in the selection of configurations in process planning.

Most MCDM approaches are based on an initial re-structuring and decomposition of the decision problem [6]. According to [6], this first step is an essential and difficult part in decision support and has significant importance for the quality of results, because all subsequent MCDM process steps are based on this re-structuring and decomposition. Furthermore, it is stated that this step receives little attention in the MCDM literature despite its importance.

In the following chapter, we will present a universal methodology for intelligent solution space exploration, which will address this core challenge of problem re-structuring and decomposition in MCDM applications.

## 3. Universal Methodology for Intelligent Solution Space Exploration

The key idea of our universal methodology for intelligent solution space exploration is to apply the *divide-and conquer-principle*. This is a prominent way to handle complex problems through an initial decomposition of the problem, followed by separately solving the resulting sub-problems and finding the paramount solution by then aggregating the partial solutions.

An *overview of our methodology* for intelligent solution space exploration is given in Fig. 1. It relies on a vertical sub-division representing different sequentially interdependent sub-problems of the paramount problem. Each vertical layer results in a number of possible solutions which lead to a horizontal sub-division representing different possible solution variants on the following vertical layer. The level of detail respective the completeness of the solutions increases with each layer. The solutions of the last layer are the final exploration result, each representing a complete solution to the paramount problem.

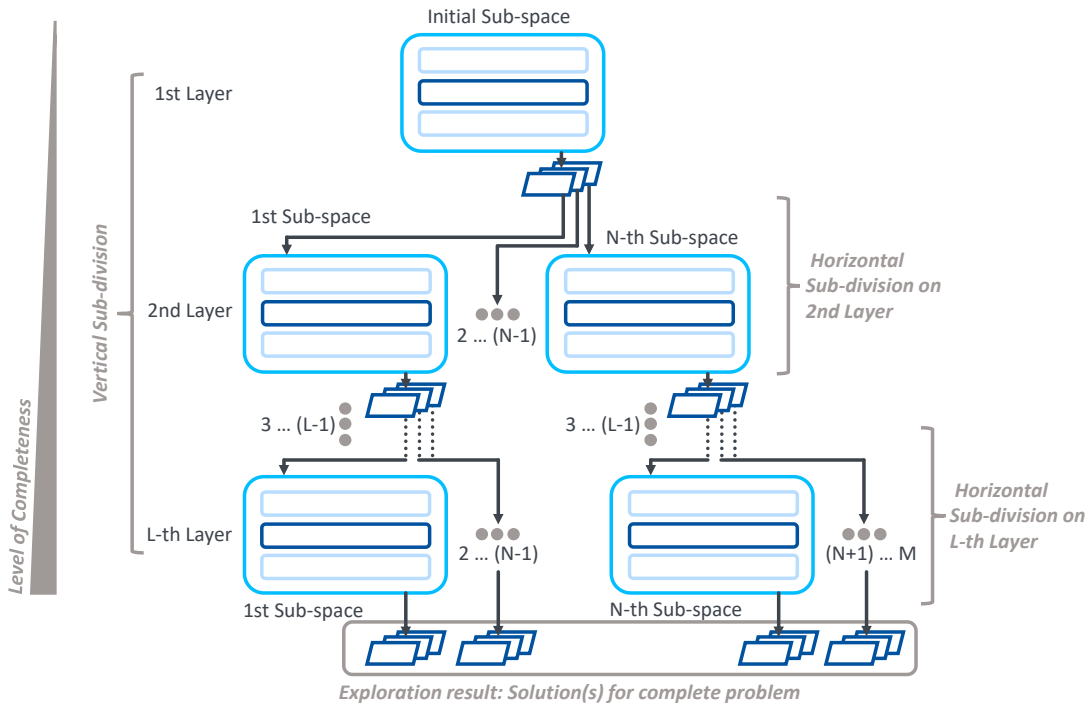

Fig. 1 Universal methodology for intelligent solution space exploration

The *vertical sub-division* aims to identify a series of connected, but sequentially solvable sub-problems, e.g. temporal phases or levels of abstraction, into which the problem can be sub-divided. It leads to between one and many vertical layers. Each layer must allow for the self-sustained solution of the respective sub-problem, but only the last layer will result in (a) solution(-s) of the paramount problem. Furthermore, each layer must be based upon a solution of the previous layer. Thus, the vertical layering can be seen as a solution completion respectively refinement process.

The solutions on each layer are subject to a filtering involving up to three filter levels each consisting of one to many filter rules (see Fig. 2). Each filter level serves a different purpose:

- The *1st filter level* can be utilized to get rid of unwanted solutions, so called "no-gos", before even exploring the respective sub-solution space. These no-gos are of non-functional nature, i.e. stemming from quality criteria, their introduction provides the possibility to limit the sub-solution space early on within a layer. The 1st filter level is optional.
- The *2nd filter level* is utilized to formulate the functional requirements regarding the respective (sub-)problem's solution. This is a mandatory task with the objective of reducing the sub-solution space so that only feasible solutions are considered further on.
- The *3rd filter level* is utilized to formulate soft criteria regarding the respective problem's solution. It allows for a ranking of the solutions found and thereby for a smart selection further reducing the (sub-)solution space. The 3rd filter level is optional.

If more than one sub-solution is found in a layer above the last layer, this leads to a further, *horizontal sub-division* into at this time equivalent, self-sustained solution sub-spaces.

The resulting structure is a tree-like sub-division of the paramount problem into different sub-problems (vertical) and different solution sub-spaces (horizontal) to the respective sub-problems.

This methodology has multiple advantages: The sub-division allows distributed and parallelized solving of the sub-problems and reduces their complexity. The different filter levels limit the growth of the solution space and thereby the consumption of time and computational resources. Furthermore, the 3rd filter level enables the adaptive limitation of the solution space, e.g. depending on the available computational power.

For any *practical implementation* of the described methodology, the following aspects should be considered: Depending on the given use-case and the layer under consideration, it has to be determined which filter levels should be used and in which manifestation they should be applied.

Furthermore, it is desirable to use the filter rules of levels 1 and 2 directly in the generation of the (sub-)solution space in order to save resources otherwise wasted on the creation of obviously infeasible solutions.

However, filters should be used with caution, because although the definition of more restrictive filter rules leads to a faster reduction of the solution space and, thus, to a faster solution space exploration, an early exclusion of solutions based upon only a partial view of the paramount problem might exclude good final solutions and, thus, deteriorate the exploration result's quality. In general, one should therefore try to filter as late as possible in order to be sure not to prematurely exclude important parts of the solution space and as early as necessary in order to reduce the computational burden. To find the optimal balance, it may be necessary to repeat the exploration process several times while analyzing its results.

## 4. Example for Intelligent Solution Space Exploration: Industrial Reconfiguration Management

The universal methodology for intelligent solution space exploration shall be evaluated on the use case of self-organized reconfiguration management. Reconfiguration management, in general, includes "*the identification of reconfiguration demand, the generation of alternative configurations, the evaluation of configurations, the selection of a new configuration and, as an optional extension, the execution of reconfiguration measures*" [12] (see Fig. 3). The corresponding methodology for self-organized reconfiguration management is visualized in Fig. 4 [12–15]. In the following, it will be used for the evaluation of the methodology proposed in this article

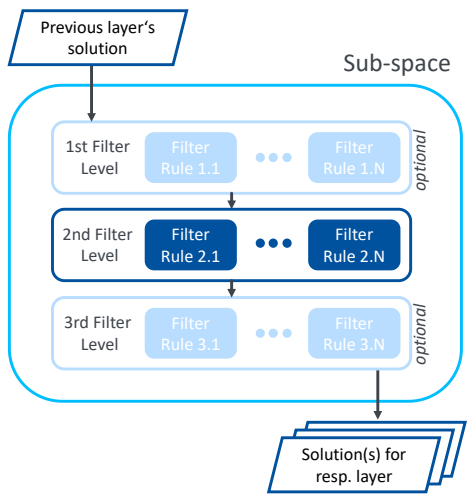

Fig. 2 Filter levels within each sub-space

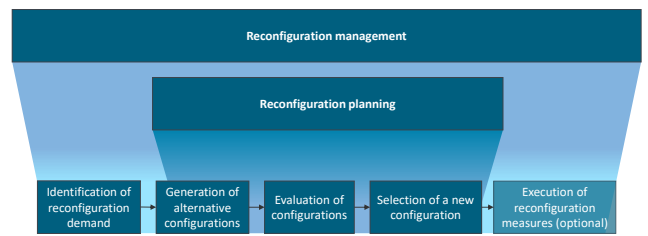

Fig. 3 Range of reconfiguration activities [15]

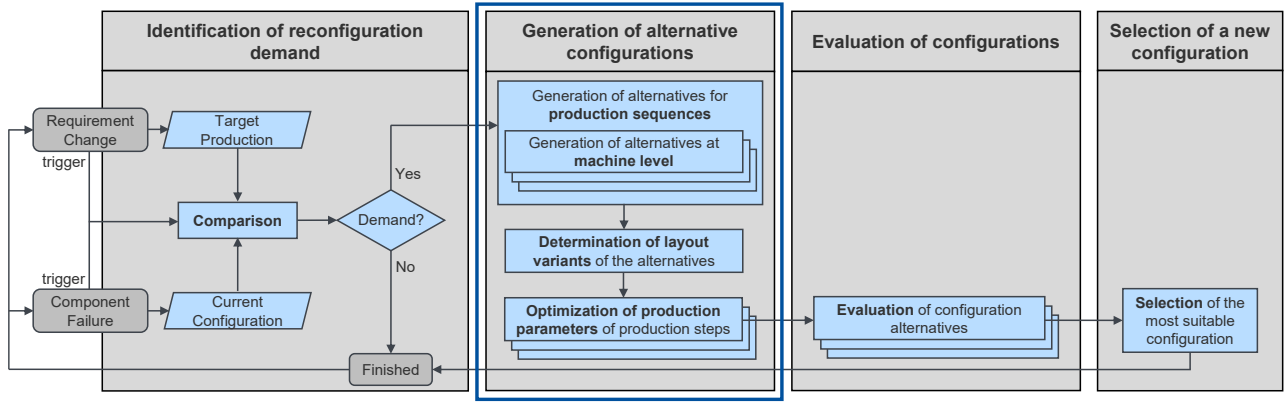

Fig. 4 Reconfiguration management methodology [12]

using an exemplary cyber-physical production system (CPPS) for discrete manufacturing.

*4.1. Cyber-physical production system (CPPS)*

The CPPS spans a modular production system (see Fig. 5), simulated in unity and controlled via a decentralized service-oriented control network. The production system possesses a matrix layout and offers discrete manufacturing services such as drilling or milling in a highly flexible and reconfigurable manner. In addition, the CPPS consists of software agents and models used to perform self-organized reconfiguration management.

Different production requirements, e.g. when a new type of product is ordered, or component failures might necessitate a reconfiguration. Once a reconfiguration demand is identified, possible solutions to fulfil the new production requirements need to be found. Afterwards, an evaluation and selection of the most suitable new configuration takes place. However, in the following, only the generation of alternative configurations will be further considered, since it is concerned with the solution space exploration.

*4.2. Generation of alternative configurations using intelligent solution space exploration*

Whenever a reconfiguration demand is identified, the solution space exploration represented in the generation of alternative configurations takes place. To do so, alternative self-organized system configurations, in our prototypical implementation represented and managed by software agents, are generated. They are refined through the three layers of the problem's *vertical sub-division*:

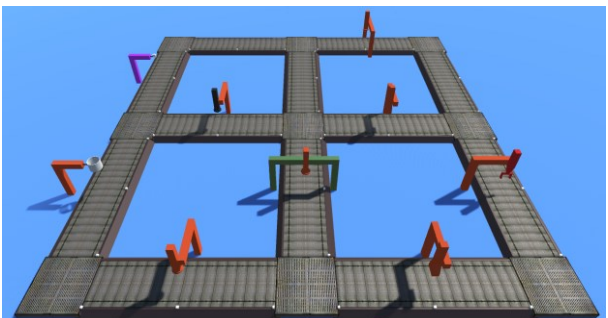

Fig. 5 Modular production system (simulation in Unity)

1. Generation of alternatives for production sequences
2. Determination of layout variants for the alternatives
3. Optimization of production parameters for the production steps

On the *first layer*, the generation of alternatives for production sequences, all feasible production sequences that can realize the required production order (target production) are determined (filter level 2, feasibility). To this end, the alternative system configurations communicate with all the cyber-physical production modules (CPPM) of the CPPS. The CPPMs determine whether they can provide a useful (manufacturing) service for the particular production sequence, in which case they are then incorporated into system configuration alternatives. The CPPMs also consider services, which they can offer in an alternative configuration (at machine level), thus they conduct the generation of alternatives at machine level. As filter level 1 (no-gos) rules, individual CPPMs and complete alternative system configurations that exceed a certain standby power consumption threshold can be excluded from further consideration. Each alternative system configuration corresponds to a solution sub-space (*horizontal sub-division* on layer 2) in which the solutions to sub-problems of vertical layers 2 and 3 will be searched for.

On the *second layer*, concerned with the determination of layout variants, a genetic algorithm is utilized to find an adjustable number of promising (using filter levels 2 and 3) layout variants for each alternative system configuration, i.e. within each solution sub-space. Different manifestations of filter level 3, e.g. aiming for an optimized transportation effort, a reduced reconfiguration effort or a compromise solution of both, can influence the solutions found. Alternatively, a brute force approach can be utilized to generate all possible (using filter level 2 only) layout variants for the pre-selected alternative system configurations. Each layout variant, incorporating the respective, previously found alternative system configuration it is based upon, corresponds to a new solution sub-space (*horizontal sub-division* on layer 3) in which the solutions to sub-problems of vertical layers 3 will be searched for.

On the *third and last layer*, the multi-objective optimization of production parameters for each derived alternative system configuration with respect to time, costs and energy using discrete event simulations are carried out. Each configuration's simulation model is generated and optimized dynamically and fully automatically. These simulation models include full replicas of the respective control logics in order to correctly reflect the behavior of the real CPPS. We have implemented different meta-heuristic-based optimization methods, but in this article only multi-objective simulated annealing on scalarized problems [12] will be used. Filter level 2 is represented by allowing only feasible parameter settings, which are obtained from the models of the individual CPPMs used in the respective system configuration. Filter level 3 is represented by the optimizations' objectives and manually set maximum iteration numbers. Thus, the optimization implements both filter levels simultaneously.

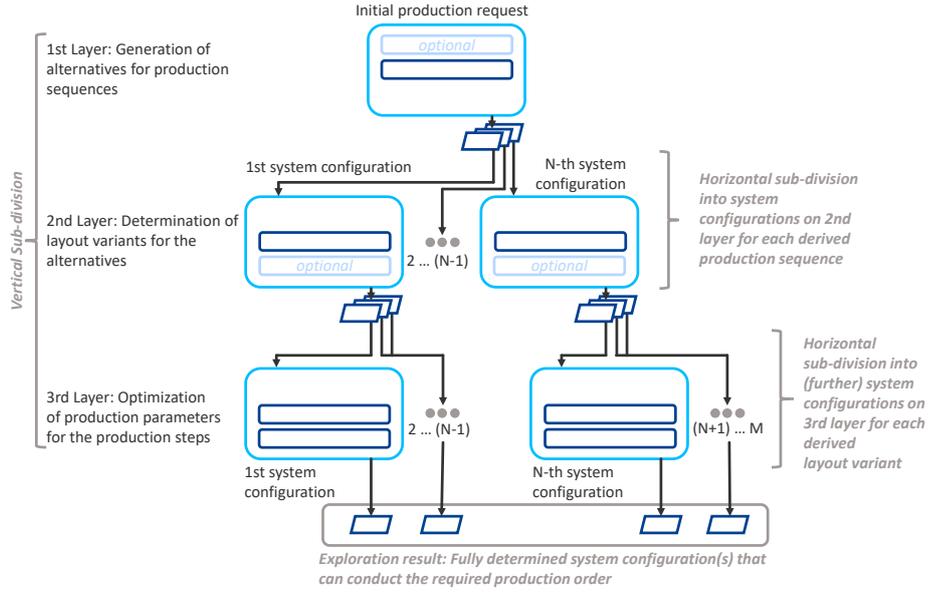

Fig. 6 Intelligent solution space exploration for the self-organized reconfiguration management use case

After the third layer, the exploration of the solution space is completed and each resulting parameter set, incorporating the respective, previously found layout variant and alternative system configuration it is based upon, corresponds to a separate final solution to the paramount problem. In the reconfiguration management process, they will then be subject to an evaluation and selection [12], which are, however, beyond the scope of this article.

### 4.3. Analysis of intelligent solution space exploration

To analyze the performance of the proposed universal methodology for intelligent solution space exploration, we compared two versions of said method with a brute force approach. An overview of the intelligent solution space exploration for the self-organized reconfiguration management, which is subsequently evaluated on a spacer disc production scenario, is given in Fig. 6. It summarizes that the

- 1st layer consist of 1st (if chosen) and 2nd level filters,
- 2nd layer consists of either 2nd and 3rd level filters or solely of 2nd level filters and the
- 3rd layer consists of 2nd and a 3rd level filters.

The two versions of the intelligent solution space exploration differ in their handling of layer 2. While version 1 (ISSEv1) does not utilize 3rd level filters on this layer, version 2 (ISSEv2) uses a genetic algorithm to find up to three most promising layout variants for which filter levels 2 and 3 are utilized. The brute force approach does not split the problem into different sub-problems; thus, more has to solve it in a single step.

Table 1 lists the results of the three different approaches to solution space exploration on the described use case. Because the setup and execution of as well as the result retrieval from the simulation models for a single system configuration takes about 1 second, the brute force approach would take an estimated time of $1.9 \cdot 10^{12}$ s, i.e. more than sixty thousand years, to generate all $1.9 \cdot 10^{12}$ possible solution variants and check them for feasibility. Thus, the brute force approach is obviously not practically applicable. Regarding ISSEv1 and ISSEv2, it can be noted that the proposed method starts to provide a benefit right from the 1st layer where it limits the number of thereafter to be considered configuration alternatives to just $p = 9$ using 2nd level filters. Layer 2, then, either results in $l = 1188$ or even $l = 20$ layout variants which need to be optimized in layer 3. The listed $t_{n,min}$ and $t_{n,max}$ values refer to the execution times in a fully parallelized or fully sequential manner. With resulting execution times $t_{tot}$ for the full solution space exploration of between about thirteen and a half minutes (for both ISSE versions) and about 4 hours (ISSEv2) respectively 214 hours (ISSEv1), the benefits of the proposed solution are easy to grasp.

Table 1: Comparison of solution space exploration methods

|  | Brute Force | ISSEv1 | ISSEv2 |
|---|---|---|---|
| L1 |  | $p = 9;\ t_p = 1\ min\ 30\ s$ | $p = 9;\ t_p = 1\ min\ 30\ s$ |
| L2 | $n = 1.9 \cdot 10^{12}$ | $l = 1188;\ t_l = 1\ min\ 11\ s$ | $l = 20;\ t_l = 1\ min\ 8\ s$ |
| L3 |  | $n = 1188$ $t_{n,min} = 10\ min\ 49\ s;\ t_{n,max} = 214\ h\ 10\ min\ 12\ s$ | $n = 20$ $t_{n,min} = 10\ min\ 49\ s;\ t_{n,max} = 3\ h\ 36\ min\ 22\ s$ |
| $t_{tot}$ | $1.9 * 10^{12}$s or 60,248a 220d 17h 47m 7s | $t_{tot,min} = 13\ min\ 30\ s$ to $t_{tot,max} = 214\ h\ 12\ min\ 53\ s$ | $t_{tot,min} = 13\ min\ 27\ s$ to $t_{tot,max} = 3\ h\ 39\ min$ |

It must be noted that, although no analytical solution could be acquired and a full numerical solution is infeasible, the quality of the solutions was sufficiently high for the described use case – and nine other scenarios carried out, but not described in this article. Furthermore, in practice, not a theoretical, yet unobtainable global optimum but an actually available good solution is what is more helpful in solving a given problem. By adjusting the filter rules for filter level 3, different trade-offs between the share of solution space fully explored and required computational resources can be realized.

## 5. Conclusion and Outlook

In this article, a universal methodology for the intelligent exploration of solution spaces is presented and evaluated on a reconfiguration management use case.

Our methodology fundamentally relies on the application of the divide-and-conquer principle, decomposing a paramount problem into sequentially interdependent sub-problems thought of as vertical layers. Each vertical layer results in a number of possible solutions leading to a horizontal sub-division representing different, self-sustained solution sub-spaces, which finally depict diverse solutions of the paramount problem. Each individual solution representative, in the evaluation scenario implemented as a software agent, evolves itself across the layers by exploring their respective solution sub-spaces. The solutions on each layer are subject to a filtering involving up to three filter levels each consisting of one to many filter rules.

Our main findings are:
- Because of the problem's decomposition, a distributed and parallelized exploration of the resulting solution sub-spaces becomes possible.
- Due to the filtering, each exploration can be tailored to the respective solution quality requirements and available computational resources.
- In our evaluation use case, the proposed method was up to 1.5 million times faster than a brute force approach (both executed sequentially) and demonstrated the ability to tailor the exploration process.

Future research should focus on a broader evaluation of the proposed method involving different scenarios possibly from different domains. A first step in this direction is already done by [16].